\def\paperlanguage{} 

\documentclass[letterpaper, 10 pt, conference]{ieeeconf}  

\usepackage{bm}
\usepackage{cite}
\usepackage{flushend}
\include{preamble}

\IEEEoverridecommandlockouts                              

\title{\LARGE \textbf
  {
    \switchlanguage%
    {%
      CubiXMusashi: Fusion of\\
      Wire-Driven CubiX and Musculoskeletal Humanoid Musashi\\
      toward Unlimited Performance
    }%
    {%
      CubiXMusashi: Fusion of\\
      Wire-Driven CubiX and Musculoskeletal Humanoid Musashi\\
      toward Unlimited Performance
    }%
  }
}

\author{Shintaro Inoue$^{1}$, Kento Kawaharazuka$^{1}$, Temma Suzuki$^{1}$, Sota Yuzaki$^{1}$, \\Yoshimoto Ribayashi$^{1}$, Yuta Sahara$^{1}$, Kei Okada$^{1}$
  \thanks{$^{1}$ The authors are with the Department of Mechano-Informatics, Graduate School of Information Science and Technology, The University of Tokyo, 7-3-1 Hongo, Bunkyo-ku, Tokyo, 113-8656, Japan.
    {\texttt\small [s-inoue, kawaharazuka, t-suzuki, yuzaki, ribayashi, sahara, k-okada]@jsk.t.u-tokyo.ac.jp}
  }
}
\begin{document}

\maketitle
\thispagestyle{empty}
\pagestyle{empty}

\begin{abstract}
  \switchlanguage%
  {%
    Humanoids exhibit a wide variety in terms of joint configuration, actuators, and degrees of freedom, 
    resulting in different achievable movements and tasks for each type. 
    Particularly, musculoskeletal humanoids are developed to closely emulate human body structure and movement functions, 
    consisting of a skeletal framework driven by numerous muscle actuators. 
    The redundant arrangement of muscles relative to the skeletal degrees of freedom 
    has been used to represent the flexible and complex body movements observed in humans.
    However, due to this flexible body and high degrees of freedom, 
    modeling, simulation, and control become extremely challenging, limiting the feasible movements and tasks.
    In this study, we integrate the musculoskeletal humanoid Musashi with the wire-driven robot CubiX, 
    capable of connecting to the environment, to form CubiXMusashi. 
    This combination addresses the shortcomings of traditional musculoskeletal humanoids 
    and enables movements beyond the capabilities of other humanoids. 
    CubiXMusashi connects to the environment with wires and drives by winding them, 
    successfully achieving movements such as pull-up, rising from a lying pose, and mid-air kicking, 
    which are difficult for Musashi alone.
    This concept demonstrates that various humanoids, not limited to musculoskeletal humanoids, 
    can mitigate their physical constraints and acquire new abilities 
    by connecting to the environment and driving through wires.
  }%
  {%

    ヒューマノイドは，関節構成やアクチュエータ，制御自由度の観点において多種多様であり，
    それぞれのヒューマノイドごとに実現可能な運動やタスクは異なる．
    特に，筋骨格ヒューマノイドは，ヒューマノイドが人間の身体構造や運動機能により近づくことを目指して開発されてきており，
    身体を構成する骨格と，それらを駆動する多くの筋肉アクチュエータからなる．
    骨格の自由度に対して冗長に配置された筋肉によって，人間に見られる柔軟で複雑な身体運動を表現してきた．
    しかし，その柔軟な身体と高い自由度がゆえにモデリングやシミュレーション，制御は困難を極め，
    実際に実現できる動きやタスクは制限されてきた．
    そこで，本研究では，筋骨格ヒューマノイドであるMusashiに，
    環境接続可能なワイヤ駆動ロボットであるCubiXを合体させてCubiXMusashiとし，
    従来の筋骨格ヒューマノイドの欠点を補いながら，
    他のヒューマノイドにも実現できない運動まで可能にする．
    CubiXMusashiが環境にワイヤを接続し，それを巻き取り駆動することで，
    Musashi単体では実現が難しい懸垂や起き上がり，キックなどの運動の実現に成功した．
    このコンセプトは，筋骨格ヒューマノイドに限らず，
    様々なヒューマノイドがワイヤを環境に接続して駆動することで，
    ヒューマノイドが持つ物理的な制約を緩め，新たな能力を獲得することを示すものである．
  }%
\end{abstract}

\section{Introduction}\label{sec:introduction}
\switchlanguage%
{%
  \begin{figure}[t]
    \begin{center}
      \includegraphics[width=0.9\columnwidth]{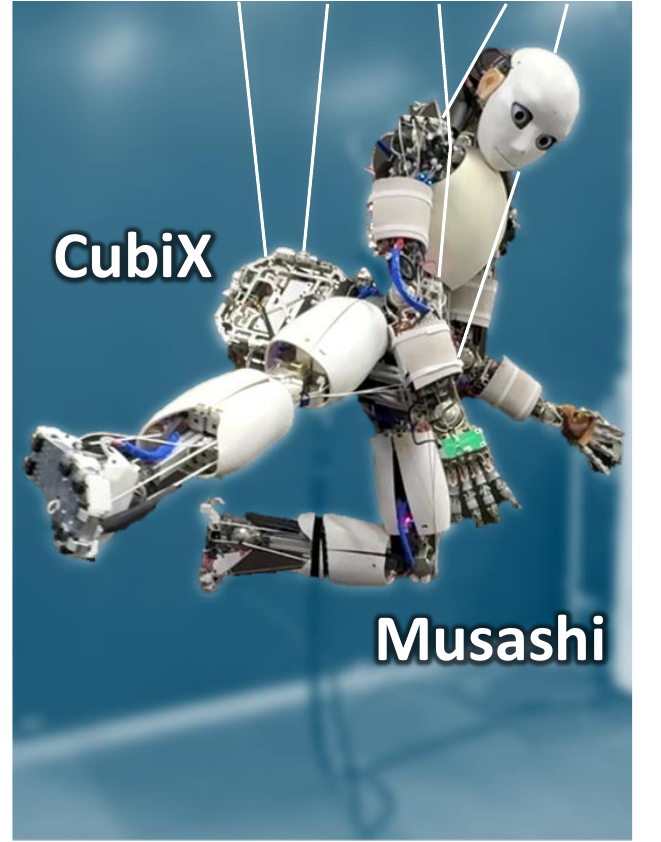}
      \vspace{-2.0ex}
      \caption{The overview of CubiXMusashi. 
      CubiXMusashi is a robot that combines Musashi, a musculoskeletal humanoid, 
      with CubiX, a wire-driven robot capable of connecting to the environment.
      }
      \vspace{-6.5ex}
      \label{fig:cubixmusashi}
    \end{center}
  \end{figure}

  In recent years, the development of humanoids has made remarkable progress, 
  leading to the creation of various full-scale humanoids such as DRC-HUBO+\cite{8281682}, 
  Digit\cite{digit}, ARTEMIS\cite{Ahn2023ARTEMIS}, and Atlas\cite{Atlas}. 
  Moreover, aiming to mimic human body structure and movement functions, 
  musculoskeletal humanoids such as ECCEROBOT\cite{diamond2017anthropomimetic}, 
  Pneumat-BS\cite{6095091}, and Anthrob\cite{jantsch2013anthrob} have been developed.

  Musculoskeletal humanoids are characterized by skeletal structures incorporating joints with high degrees of freedom, 
  such as ball joints and universal joints, which are driven by numerous actuators resembling muscles.
  Leveraging these flexible bodies and high degrees of freedom, 
  the humanoid has been able to perform flexible and complex movements, such as driving a car, 
  similar to those performed by humans\cite{kawaharazuka2020driving}. 
  However, the flexibility of their bodies and high degrees of freedom
  have made precise modeling and simulation of the entire robot body extremely difficult. 
  As a result, precise and responsive control is challenging, and the range of achievable movements and tasks has been limited.

  Focusing on robots driven by wires connected to the environment, 
  it has been demonstrated that connecting a wire extending from a vehicle to the environment using a drone, 
  and then the vehicle winding the wire, improves the vehicle's mobility\cite{8794265}.
  Furthermore, studies have been conducted where humanoid robots utilize wires connected to the environment, 
  such as HRP-2\cite{1307969} gripping a rope connected to the environment while climbing a 40-degree steep slope\cite{8594292}, 
  and descending from a height of 2.5 m along a wall\cite{8239547}.
  Additionally, 
  there is research involving attaching a winch capable of winding wires to the cart-type musculoskeletal humanoid Musashi-W\cite{10000123}, 
  enabling assistance in tasks such as lifting and pulling heavy objects\cite{10375200}.
  These examples show 
  that using wires connected to the environment enables augmentation and expansion of humanoid motion.

  Therefore, this study integrates Musashi, a musculoskeletal humanoid, with CubiX\cite{inoue2024cubix}, 
  an environment-connectable wire-driven robot, to form CubiXMusashi, as shown in \figref{fig:cubixmusashi}. 
  This integration aims to achieve motion capabilities, 
  which were not achievable with traditional musculoskeletal humanoids. 
  This approach not only enhances the motion capabilities of musculoskeletal humanoids 
  but also proposes that by using wires connected to the environment, 
  humanoids can mitigate their physical constraints and acquire new motion capabilities. 
  In this paper, we emphasize 
  introducing the concept of aiming to achieve unlimited performance with humanoid robots 
  equipped with the ability to connect wires to the environment and utilize them for operation, 
  and do not delve extensively into methods on control theory and motion planning.
}%
{%
  近年のヒューマノイド開発は目覚ましく，
  DRC-HUBO+\cite{8281682}，Digit\cite{digit}，ARTEMIS\cite{Ahn2023ARTEMIS}，Atlas\cite{Atlas}などの
  多種多様な等身大ヒューマノイドが開発されてきた．
  また，ヒューマノイドが人間の身体構造や運動機能により近づくことを目指して，
  筋骨格ヒューマノイドであるECCEROBOT\cite{gravato2010ecce1, diamond2017anthropomimetic}や，Peneumat-BS\cite{6095091}，
  Anthrob\cite{jantsch2013anthrob}などが開発されてきた．

  筋骨格ヒューマノイドは，
  球関節や自在継手などの高い自由度を持つ関節が多く取り入れられた身体構造である骨格を，
  筋肉を模した多くのアクチュエータによって駆動する．
  例えば，筋骨格ヒューマノイドであるMusashi\cite{kawaharazuka2019musashi}は，
  手を除いて32自由度ある身体を，74個の筋肉によって駆動する．
  筋肉は，ワイヤを巻き取り張力を発生させる筋モジュールで表現されている．
  これらの柔軟な身体や高い制御自由度を活かして，
  人間が行う車の運転などの柔軟で複雑な動きを表現してきた\cite{kawaharazuka2020driving}．
  一方で，柔軟な身体や高い制御自由度，摩擦や伸びが発生するワイヤによる駆動が要因となり，
  ロボット全身の正確なモデリングやシミュレーションは非常に困難である．
  そのため，精密かつ即応性のある制御は難しく，実際に実現可能である動作やタスクは限られてきた．

  \begin{figure}[t]
    \begin{center}
      \includegraphics[width=1\columnwidth]{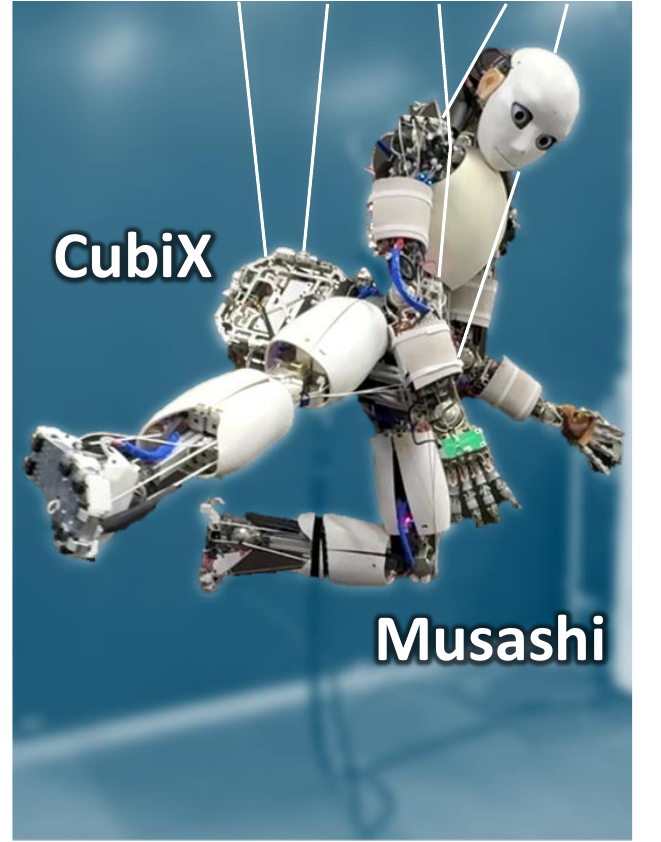}
      \vspace{-4.0ex}
      \caption{The overview of CubiXMusashi. 
      CubiXMusashi is a robot that combines Musashi, a musculoskeletal humanoid, 
      with CubiX, a wire-driven robot capable of connecting to the environment.
      By extending wires from its own body and connecting them to the environment, 
      the humanoid achieves a level of motion performance that was previously difficult to realize.
      }
      \vspace{-6.0ex}
      \label{fig:cubixmusashi}
    \end{center}
  \end{figure}

  ここで，環境に接続されたワイヤを用いて駆動するロボットに注目すると，
  車両から伸びるワイヤをドローンが環境に接続し，そのワイヤを車両が巻き取ることで，
  車両の走破性が向上することを示す研究\cite{8794265}がある．
  さらに，ヒューマノイドが環境に接続されたワイヤを利用する研究として，
  HRP-2\cite{1307969}が環境に接続されたロープを掴みながら，40度の急斜面を登る研究\cite{8594292}や，
  2.5 mの高所から壁を伝って降りる研究\cite{8239547}がある．
  また，台車型筋骨格ヒューマノイドMusashi-W\cite{10000123}に
  ワイヤを巻き取ることができるウインチを取り付け，
  カラビナを用いてそのワイヤを環境に接続して用いる研究では，
  重量物の持ち上げや牽引など，ヒューマノイドにおける発揮力の補助を実現した\cite{10375200}．
  このように，環境に接続したワイヤを用いることでヒューマノイドの運動の補助，拡張が可能であることがわかる．

  そこで，本研究では，筋骨格ヒューマノイドであるMusashiに，
  環境接続可能なワイヤ駆動ロボットであるCubiX\cite{inoue2024cubix}を合体させ，
  新たに\figref{fig:cubixmusashi}に示されるCubiXMusashiとし，
  従来の筋骨格ヒューマノイドでは実現されなかった懸垂動作，起き上がり動作，キック動作などの運動性能を実現する．
  これにより，
  筋骨格ヒューマノイドの運動性能を向上させることに加えて，
  一般にヒューマノイドが環境に接続したワイヤを用いて駆動することで，
  ロボットが持つ物理的な制約を緩め，新たに運動機能を獲得することを提案する．
  なお，本論文は環境利用ワイヤ駆動を兼ね備えたヒューマノイドによる
  制限のない性能を目指すコンセプトの紹介を重視し，制御理論や動作計画について多くは語らない．



}%

\section{Methods} \label{sec:methods}
\subsection{Overall Structure of CubiXMusashi}
\switchlanguage%
{%
  \begin{figure}[t]
    \centering
    \includegraphics[width=1.0\columnwidth]{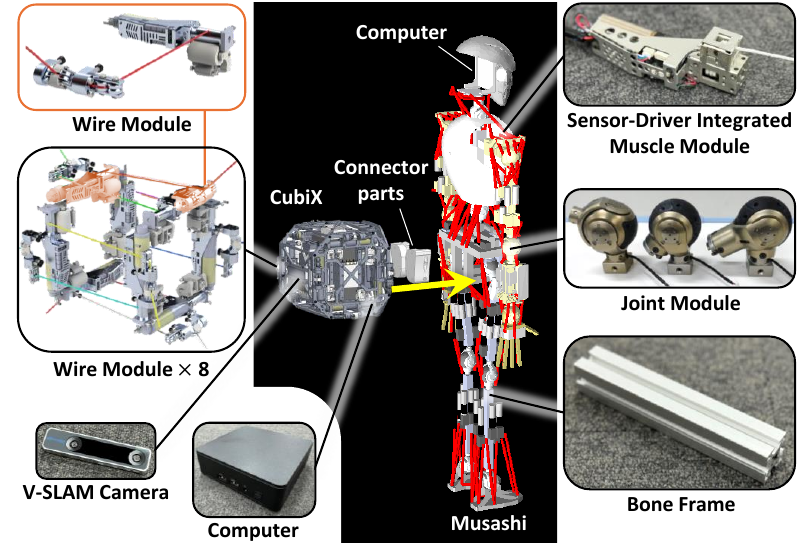}
    \vspace{-4ex}
    \caption{
      The hardware structure of CubiXMusashi.
      CubiX and Musashi were assembled on the back pelvic area of Musashi using connector parts. 
      CubiX is equipped with 8 wire modules, sensors, and a computer. 
      Musashi's body consists of muscle modules representing muscles, joint modules functioning as joints, 
      and bone frames serving as the skeleton.
    }
    \vspace{-4ex}
    \label{fig:hardware}
  \end{figure}

  The hardware configuration of CubiXMusashi is shown in \figref{fig:hardware}.
  CubiXMusashi is a robot that combines the wire-driven robot CubiX, which is capable of environment connection, 
  with the musculoskeletal humanoid Musashi using connector parts. 
  It stands 1.6 m tall and weighs 44.6 kg excluding batteries.

  Musashi is a full-scale musculoskeletal humanoid consisting of muscle actuators, joints, and skeleton. 
  Muscle actuators are represented by muscle modules that generate tension by winding wires. 
  Each muscle module integrates a motor for wire winding, a temperature sensor to monitor motor temperature, 
  a load cell to measure wire tension, and a motor driver for control and integration. 
  With 74 muscle modules corresponding to the number of muscles, they collectively drive the body. 
  Ball-shaped joint modules are used at the joints.
  Moreover, by combining modular components, the degrees of freedom of joints can be altered, 
  enabling the construction of human-like joint except for the scapulohumeral joint. 
  A universal bone frame is used for the skeleton, facilitating assembly of muscle modules and connection to joint modules.
  In this study, CubiX is assembled onto Musashi without interference with Musashi's muscles, 
  attached to the exposed pelvis on the back of the skeleton, taking advantage of Musashi's universal bone frame construction.

  CubiX embeds actuators that wind wires inside its body. 
  It drives by connecting multiple wires extending from the body to the environment and winding them for operation.
  8 wire modules are assembled along the edges of the cube-shaped body structure. 
  Each wire module has a maximum continuous tension of 180 N.
  Furthermore, leveraging its cube-shaped body structure, it is possible to integrate tools or other robots into CubiX.
  This study also exploits this feature, achieving integration between CubiX and the musculoskeletal humanoid Musashi.
  CubiX's body integrates a computer and V-SLAM camera, enabling autonomous control and localization solely by CubiX. 
  While autonomous wire connection to the environment using drones is feasible, 
  this aspect is omitted and not addressed in this study.

  \begin{figure}[t]
    \centering
    \includegraphics[width=1.0\columnwidth]{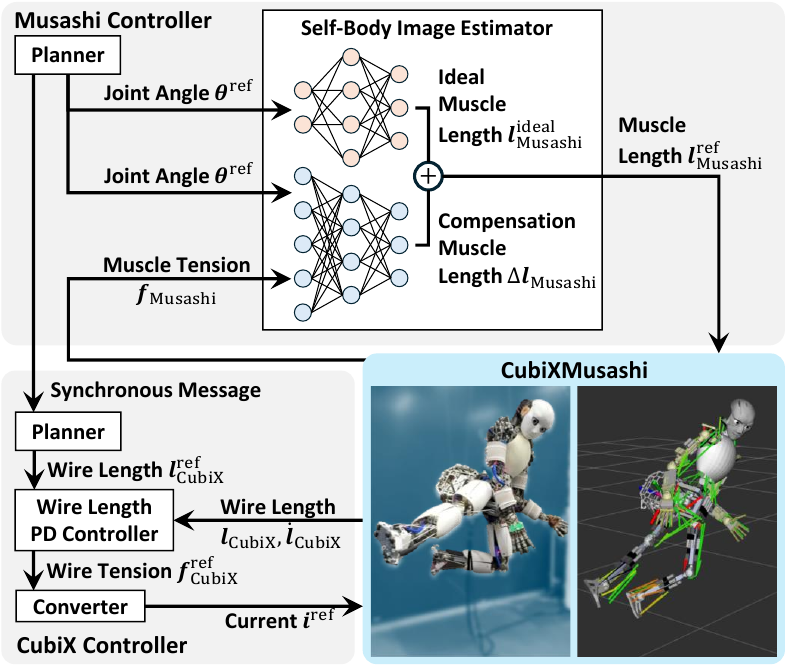}
    \vspace{-4ex}
    \caption{
      The system configuration of CubiXMusashi.
      Separate controllers run for Musashi and CubiX, 
      and CubiXMusashi is controlled by synchronizing the planners in each controller. 
      Musashi was controlled using inference by neural networks, and CubiX used wire length control.
    }
    \vspace{-4ex}
    \label{fig:system}
  \end{figure}

  \begin{figure*}[t]
    \centering
    \includegraphics[width=2.02\columnwidth]{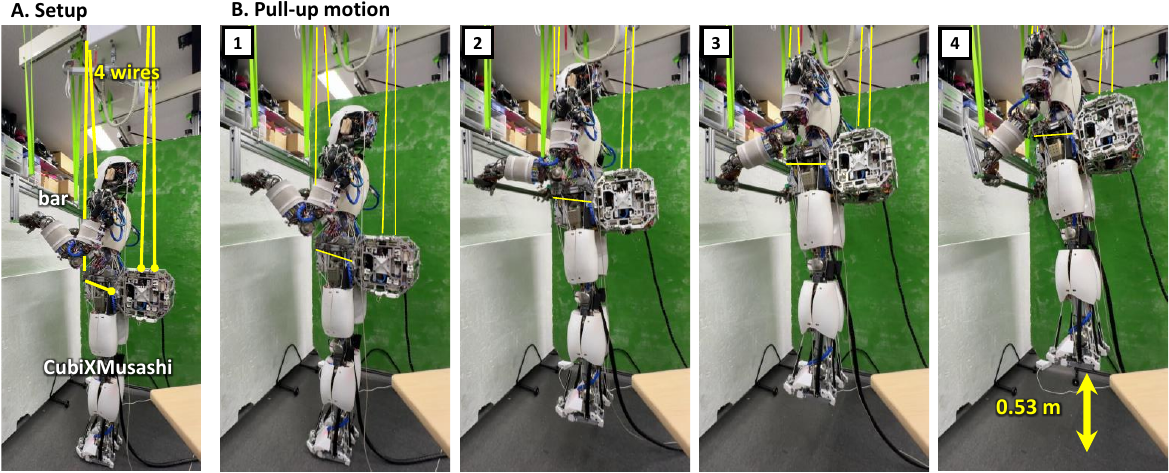}
    \vspace{-1.5ex}
    \caption{
      Pull-up motion experiment.
      (A) Experimental setup.
      (B) Pull-up motion.
      By winding the 4 wires connected to the environment, 
      it was possible to perform a pull-up motion, lifting the entire body.
    }
    \vspace{-4ex}
    \label{fig:exp1_pic}
  \end{figure*}

  \begin{figure}[t]
    \centering
    \includegraphics[width=0.9\columnwidth]{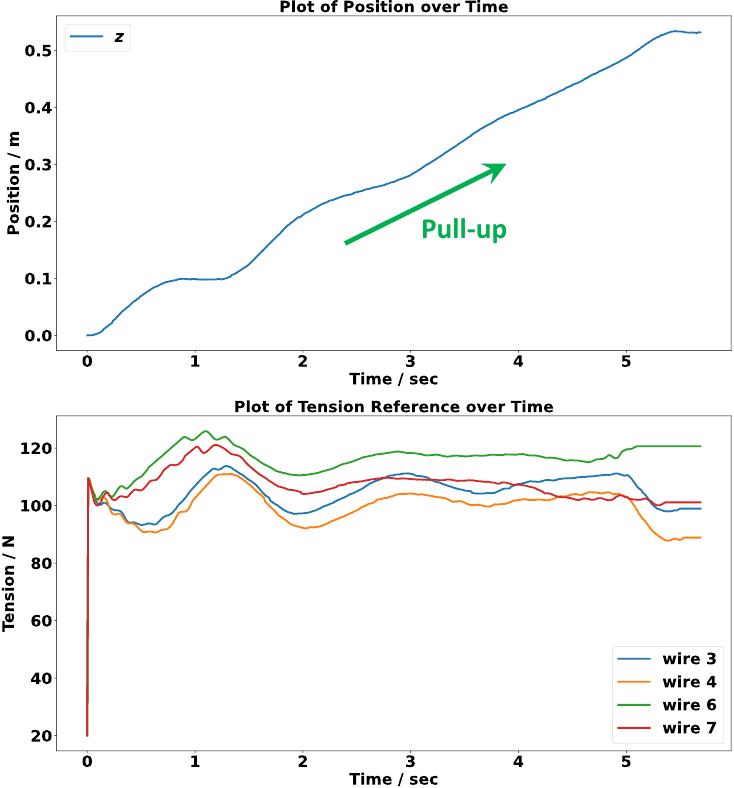}
    \vspace{-2ex}
    \caption{
      Time series data of height position $z$ and target wire tension $\bm{f}^\mathrm{ref}_\mathrm{CubiX}$ during the pull-up motion experiment. 
      By controlling the length of the 4 wires, 
      it shows that the entire body of CubiXMusashi was lifted while compensating for its own weight.
    }
    \vspace{-5ex}
    \label{fig:exp1_data}
  \end{figure}

  \begin{figure*}[t]
    \centering
    \includegraphics[width=2.02\columnwidth]{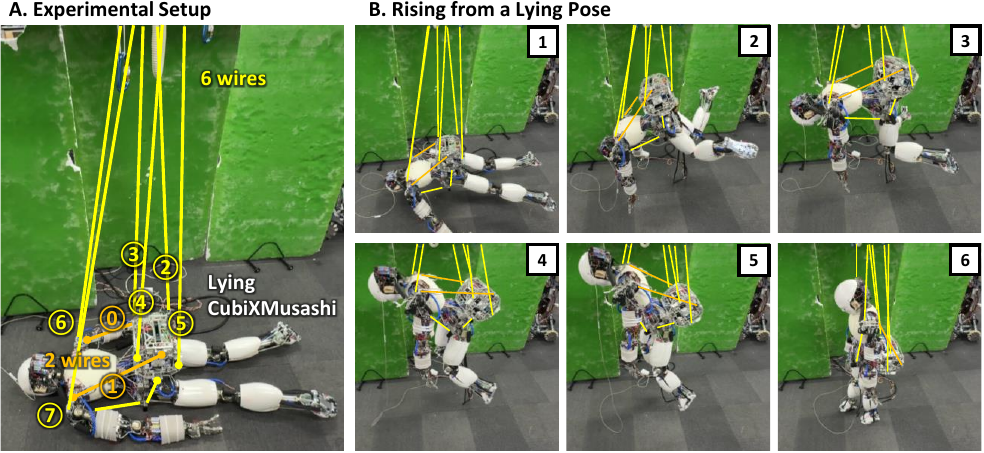}
    \vspace{-1.5ex}
    \caption{
      Rising from a lying pose experiment.
      (A) Experimental setup.
      (B) Rising from a lying pose.
      Using 6 wires connected to the environment and 2 wires connected to its own body,
      it shows that the robot achieves a rising motion, 
      transitioning to a standing pose by lifting its body and rotating in mid-air.
    }
    \vspace{-4ex}
    \label{fig:exp2_pic}
  \end{figure*}

  \begin{figure}[t]
    \centering
    \includegraphics[width=0.9\columnwidth]{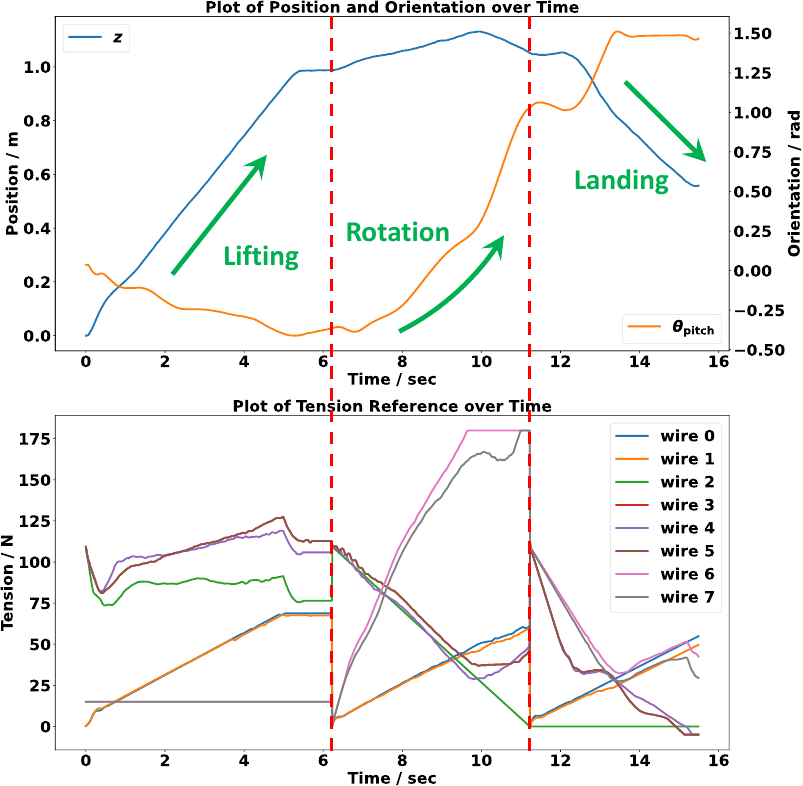}
    \vspace{-2ex}
    \caption{
      Time series data of height position $z$, pitch direction rotation angle $\theta_\mathrm{pitch}$, 
      and target wire tension $\bm{f}^\mathrm{ref}_\mathrm{CubiX}$ during the rising from a lying pose experiment. 
      By controlling the length of 8 wires, 
      it shows that the robot sequentially performed the ascent phase, 
      rotation phase, and landing phase, achieving the rising motion.
    }
    \vspace{-4.5ex}
    \label{fig:exp2_data}
  \end{figure}

  }%
{%
  \begin{figure}[t]
    \centering
    \includegraphics[width=1.0\columnwidth]{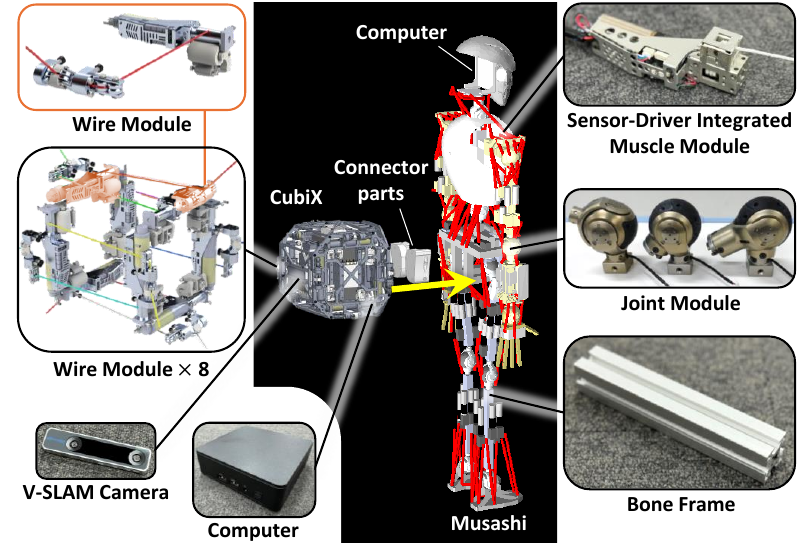}
    \vspace{-4ex}
    \caption{
      The hardware structure of CubiXMusashi.
      CubiX and Musashi were assembled on the back pelvic area of Musashi using connector parts. 
      CubiX is equipped with 8 wire modules, sensors, and a computer. 
      Musashi's body consists of muscle modules representing muscles, joint modules functioning as joints, 
      and bone frames serving as the skeleton.
    }
    \vspace{-3ex}
    \label{fig:hardware}
  \end{figure}
  CubiXMusashiのハードウェア構成を\figref{fig:hardware}に示す．
  CubiXMusashiは，環境接続可能なワイヤ駆動ロボットであるCubiXを，
  筋骨格ヒューマノイドであるMusashiにコネクタパーツを用いて組み付けたロボットである．
  身長は1.6 m，体重はバッテリを除いて44.7 kgである．

  Musashiは，
  全身を筋肉アクチュエータと関節，骨格で構成する等身大の筋骨格ヒューマノイドである．
  筋肉アクチュエータは，ワイヤを巻き取り張力を発生させる機能を持つ筋モジュールで表現されている．
  筋モジュールには，ワイヤを巻き取るモータ，そのモータ温度を測る温度センサ，ワイヤ張力を計測するロードセル，
  それらの制御，統合を行うモータドライバが集約されている．
  この筋モジュールを筋肉の数だけ用意することで，計74筋で身体を駆動している．
  関節には，球形の関節モジュールが用いられ，ポテンショメータとIMUを内蔵することにより，正確な姿勢の測定を可能にしている．
  また，関節モジュールの再配置可能な部品を組み合わせることで，関節の自由度を変更することが可能であり，
  肩甲骨関節を除いて簡略化された人間の関節自由度を構築できる．
  骨格には，汎用的な骨フレームが用いられており，筋モジュールの組み付けや，関節モジュールへの接続などを容易としている．
  本研究において，CubiXのMusashiへの組み付けは，Musashiの筋肉との干渉がなく，
  骨格がむき出しになっている背中側の骨盤に行った．
  これはMusashiの骨格が上記の汎用的な骨フレームで構成されている利点を活かして実現されている．

  CubiXは，ワイヤを巻き取るアクチュエータをロボット体内に内蔵し，
  ロボット体内から出る複数のワイヤを環境に接続させ，それを巻き取ることで駆動するロボットである．
  キューブ形状の身体構造の辺に沿ってワイヤを巻き取るワイヤモジュールが合計で8個，組み付けられている．
  ワイヤモジュールは最大連続張力 18 kgf，ワイヤ巻取り速度 242 mm/s, ワイヤ巻取り長さ 5.3 m という性能を持ち，
  これは，3本以上のワイヤを鉛直上向きに環境に接続すれば，
  CubiXMusashiの全重量を補償しながらワイヤ長さを制御するのに十分な出力である．
  さらに，キューブ形状である身体構造を活かして，CubiXに道具や他のロボットを合体させることが可能であり，
  本研究でもその特徴を活かして，筋骨格ヒューマノイドであるMusashiとの合体を果たしている．
  CubiXの体内にはコンピュータやV-SLAMカメラを内蔵しており，自身の制御，自己位置推定がCubiXだけで完結するようになっている．
  また，環境へのワイヤの接続は，ドローンを用いて自律的にすることも可能だが，本研究では省略し，扱わないこととする．
}%

\subsection{System Configuration and Controller of CubiXMusashi}
\switchlanguage%
{%
  The system configuration of CubiXMusashi is shown in \figref{fig:system}. 
  CubiXMusashi is a robot that combines 2 robots into one, but they have separate power systems and computers. 
  Regarding robot control, each robot is controlled by its respective computer, 
  and their synchronization is achieved through communication between the 2 computers 
  to generate the overall movement of CubiXMusashi.

  In controlling Musashi, 
  whole-body control is performed using Self-Body Image\cite{kawaharazuka2018bodyimage, kawaharazuka2019longtime}, 
  employing inference by the following 2 neural networks:
  \begin{enumerate}
    \item A neural network that takes the target joint angles $\bm{\theta}^\mathrm{ref}$ 
      provided by the planner as input and outputs the ideal muscle lengths $\bm{l}^\mathrm{ideal}_\mathrm{Musashi}$.
    \item A neural network that takes $\bm{\theta}^\mathrm{ref}$ 
      and the muscle tensions $\bm{f}_\mathrm{Musashi}$ measured by each muscle module as input, 
      and outputs compensatory muscle lengths $\Delta\bm{l}_\mathrm{Musashi}$ 
      to make the muscle lengths $\bm{l}^\mathrm{ideal}_\mathrm{Musashi}$ applicable in the actual robot.
  \end{enumerate}
  The outputs of these 2 neural networks are combined 
  to compute the target muscle lengths $\bm{l}^\mathrm{ref}_\mathrm{Musashi}$.
  This $\bm{l}^\mathrm{ref}_\mathrm{Musashi}$ is then inputted into the motor drivers of each muscle module 
  and the length control is performed by the FPGA mounted on the board.
  In this way, each muscle module drives to achieve the $\bm{\theta}^\mathrm{ref}$ requested by the planner, 
  which is manually designed for each experiment.

  In controlling CubiX, a PD control of wire length is performed to approach each wire length $\bm{l_\mathrm{CubiX}}$ 
  calculated from motor rotation angles to the target wire length $\bm{l}^\mathrm{ref}_\mathrm{CubiX}$ requested by the planner, 
  and to approach each wire velocity $\dot{\bm{l}}_\mathrm{CubiX}$ calculated from motor rotation speeds to 0. 
  A feedforward term to compensate for the weight of CubiXMusashi 
  is added to the target wire tension $\bm{f}^\mathrm{ref}_\mathrm{CubiX}$, which is the output of the PD control.
  This $\bm{f}^\mathrm{ref}_\mathrm{CubiX}$ is converted into command currents $\bm{i}^\mathrm{ref}$ 
  for each motor driver 
  based on the mechanical model.
  Additionally, 
  CubiX's planner synchronizes its operation timing with Musashi's planner by receiving synchronization messages, 
  ensuring coordination with Musashi's movements.
}%
{%
  \begin{figure}[t]
    \centering
    \includegraphics[width=1.0\columnwidth]{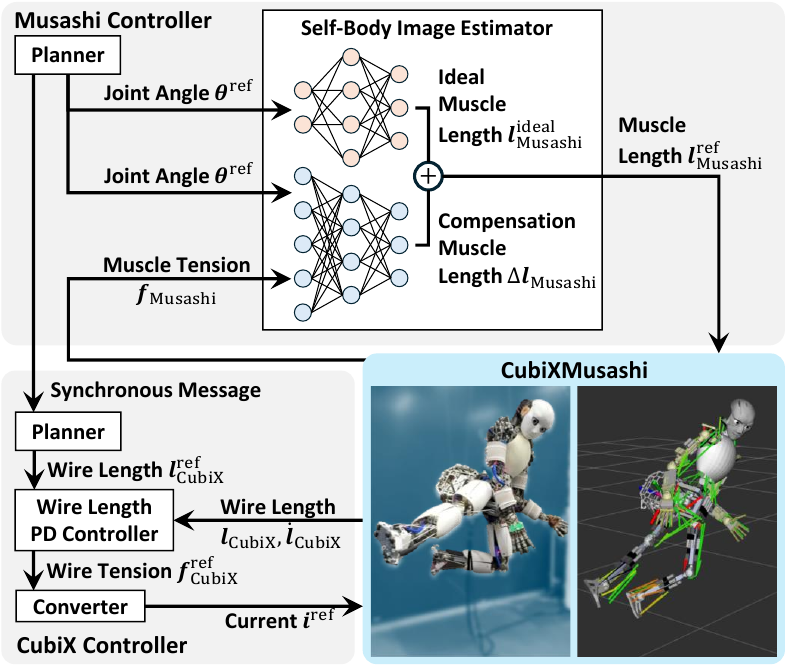}
    \vspace{-4ex}
    \caption{
      The system configuration of CubiXMusashi.
      Separate controllers run for Musashi and CubiX, 
      and CubiXMusashi is controlled by synchronizing the planners in each controller. 
      Musashi was controlled using inference by neural networks, and CubiX used wire length control.
    }
    \vspace{-3ex}
    \label{fig:system}
  \end{figure}

  \begin{figure*}[t]
    \centering
    \includegraphics[width=2.02\columnwidth]{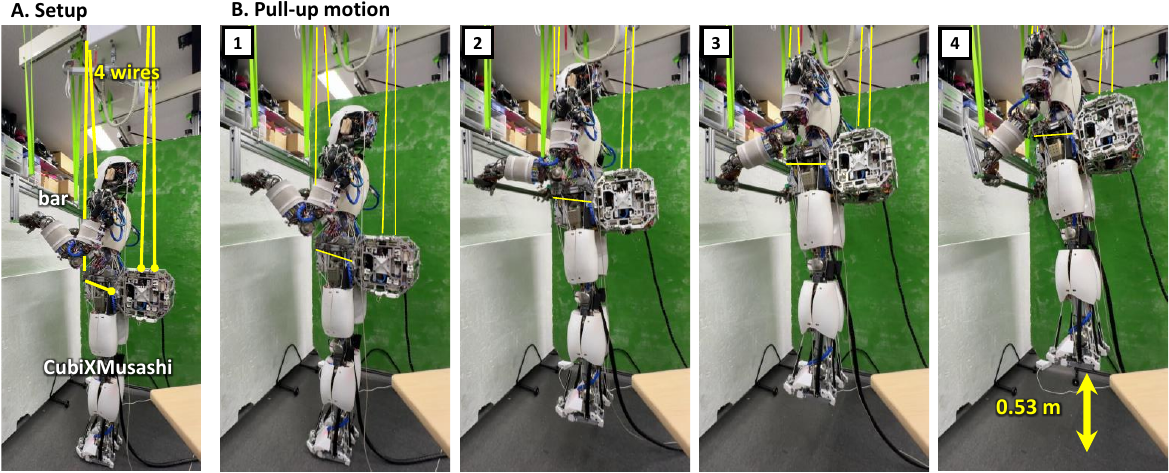}
    \vspace{-1.5ex}
    \caption{
      Pull-up motion experiment.
      (A) Experimental setup.
      (B) Pull-up motion.
      By winding the four wires connected to the environment, 
      it was possible to perform a pull-up motion, lifting the entire body.
    }
    \vspace{-3ex}
    \label{fig:exp1_pic}
  \end{figure*}

  \begin{figure}[t]
    \centering
    \includegraphics[width=1.0\columnwidth]{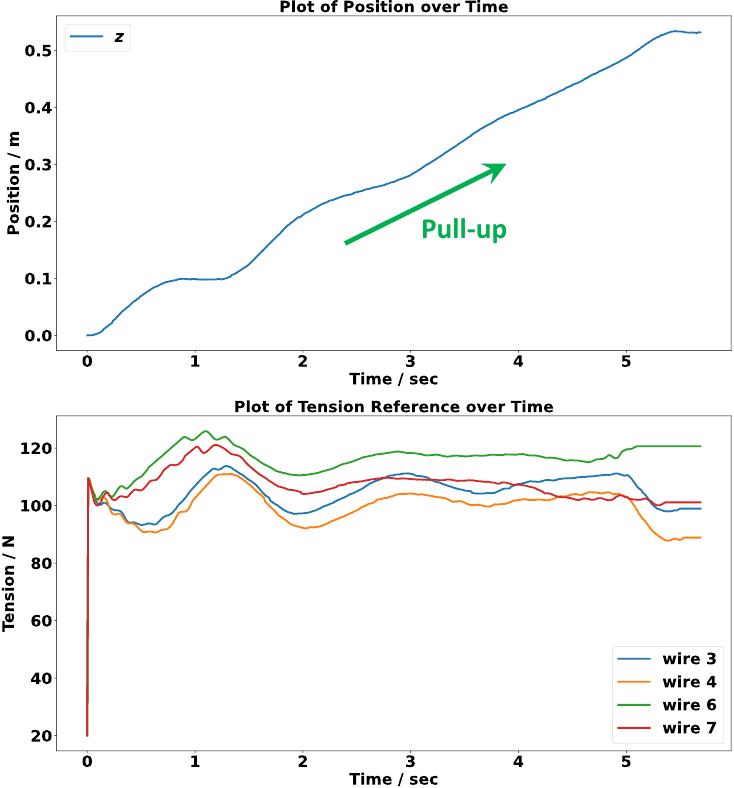}
    \vspace{-4ex}
    \caption{
      Time series data of height position $z$ and target wire tension $\bm{f}^\mathrm{ref}_\mathrm{CubiX}$ during the pull-up motion experiment. 
      By controlling the length of the 4 wires, 
      it shows that the entire body of CubiXMusashi was lifted while compensating for its own weight.
    }
    \vspace{-5ex}
    \label{fig:exp1_data}
  \end{figure}

  CubiXMusashiのシステム構成を\figref{fig:system}に示す．
  CubiXMusashiは，2台のロボットを合体させて1台のロボットとしているが，電源系統やコンピュータは別々のものになっている．
  ロボットの制御に関しても，各ロボットにあるコンピュータで各ロボットを制御するが，
  それらの同期を2台のコンピュータ間の通信によって行うことでCubiXMusashi全体の動きを生成している．

  Musashiの制御では，Self-Body Imageを用いた全身の制御\cite{kawaharazuka2018bodyimage, kawaharazuka2019longtime}が行われており，
  下記，2つのニューラルネットワークによる推論が行われている．
  \begin{enumerate}
    \item Plannerから与えられる全身の目標関節角度$\bm{\theta}^\mathrm{ref}$を入力とし，
      理想的な各筋長$\bm{l}^\mathrm{ideal}_\mathrm{Musashi}$を出力するニューラルネットワーク．
    \item 上記$\bm{\theta}^\mathrm{ref}$と各筋モジュールで測定される筋張力$\bm{f}_\mathrm{Musashi}$を入力とし，
      筋長$\bm{l}^\mathrm{ideal}_\mathrm{Musashi}$を実機で適用可能なものとするための
      補償的な筋長$\Delta\bm{l}_\mathrm{Musashi}$を出力とするニューラルネットワーク．
  \end{enumerate}
  これら2つのニューラルネットワークの出力を合わせて，目標筋長$\bm{l}^\mathrm{ref}_\mathrm{Musashi}$が計算される．
  この$\bm{l}^\mathrm{ref}_\mathrm{Musashi}$が，各筋モジュールのモータドライバに入力され，
  基板に搭載されたFPGAにより長さ制御が行われる．
  このようにして，Plannerが要求する$\bm{\theta}^\mathrm{ref}$が達成されるように各筋モジュールが駆動する．
  ただし，この$\bm{\theta}^\mathrm{ref}$は，実験ごとに手作業で設計される姿勢である．

  CubiXの制御では，モータ回転角度から算出される各ワイヤ長さ$\bm{l_\mathrm{CubiX}}$を
  Plannerが要求する目標ワイヤ長さ$\bm{l}^\mathrm{ref}_\mathrm{CubiX}$に近づけ，
  モータ回転角速度から算出される各ワイヤ速度$\dot{\bm{l}_\mathrm{CubiX}}$を0に近づけるようなワイヤ長さPD制御を行っている．
  PD制御の出力である目標ワイヤ張力$\bm{f}^\mathrm{ref}_\mathrm{CubiX}$に
  CubiXMusashiの自重を補償する張力がフィードフォワード項として加算される．
  このように計算された$\bm{f}^\mathrm{ref}_\mathrm{CubiX}$は，
  ワイヤ巻取りプーリ半径やモータのトルク定数などのパラメタを用いて力学モデルに基づき定数倍されることで，
  各モータドライバへの司令電流$\bm{i}^\mathrm{ref}$に変換され，各ワイヤモジュールに入力される．
  なお，CubiXのPlannerはMusashiのPlannerから動作開始タイミングに同期メッセージを受け取ることで，
  Musashiの動作との連携を取っている．
}%

\section{Experiments} \label{sec:experiments}
\switchlanguage%
{%
  In this study, we conducted 3 types of experiments to show the motion capabilities of CubiXMusashi: 
  a pull-up motion experiment, a rising from a lying pose experiment, and a mid-air kicking experiment. 
  Through these experiments, 
  we demonstrate that by integrating Musashi with CubiX and driving it using wires connected to the environment, 
  movements that were not achievable with conventional musculoskeletal humanoids can be realized.
}%
{%
  \begin{figure*}[t]
    \centering
    \includegraphics[width=2.02\columnwidth]{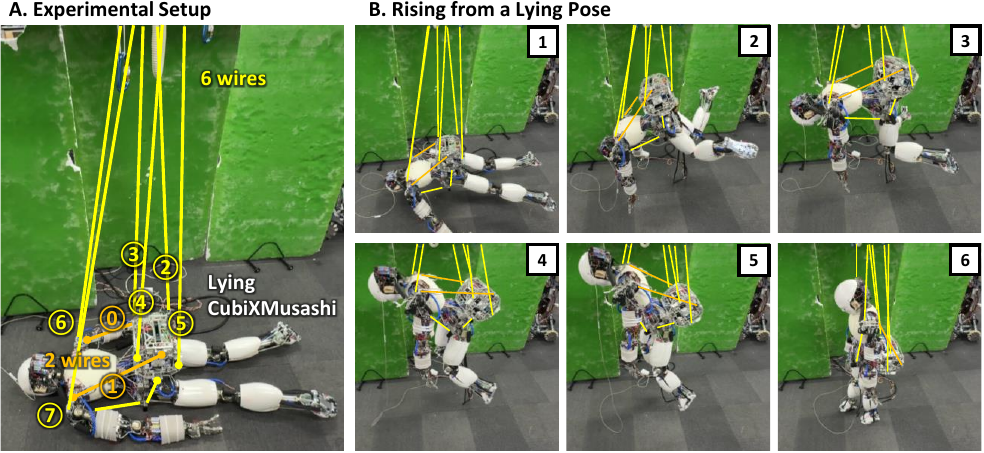}
    \vspace{-1.5ex}
    \caption{
      Rising from a lying pose experiment.
      (A) Experimental setup.
      (B) Rising from a lying pose.
      Using 6 wires connected to the environment and 2 wires connected to its own body,
      it shows that the robot achieves a rising motion, 
      transitioning to a standing pose by lifting its body and rotating in mid-air.
    }
    \vspace{-2ex}
    \label{fig:exp2_pic}
  \end{figure*}

  \begin{figure}[t]
    \centering
    \includegraphics[width=1.0\columnwidth]{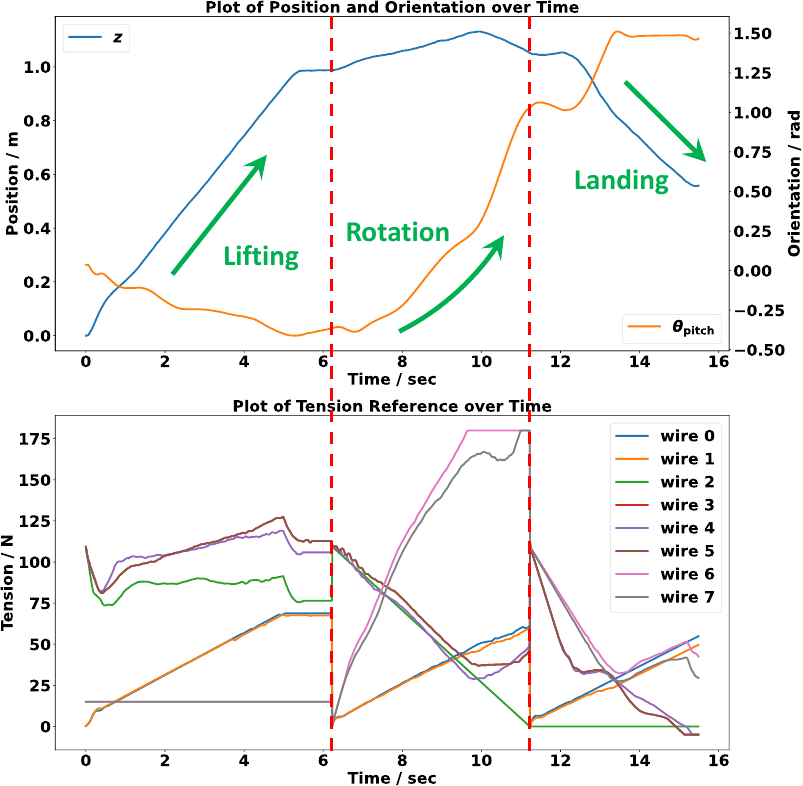}
    \vspace{-4ex}
    \caption{
      Time series data of height position $z$, pitch direction rotation angle $\theta_\mathrm{pitch}$, 
      and target wire tension $\bm{f}^\mathrm{ref}_\mathrm{CubiX}$ during the rising from a lying motion experiment. 
      By controlling the length of 8 wires, 
      it shows that the robot sequentially performed the ascent phase, 
      rotation phase, and landing phase, achieving the rising motion.
    }
    \vspace{-4.9ex}
    \label{fig:exp2_data}
  \end{figure}

  本研究では，CubiXMusashiの運動性能を示すため，
  懸垂動作実験，起き上がり動作実験，キック動作実験の3種類の実験を行った．
  懸垂動作実験では，CubiXMusashiが胸の高さにある懸垂棒を掴み，腰がその高さに持ち上がる懸垂を行う．
  起き上がり動作実験では，CubiXMusashiがうつ伏せで地面に横たわった状態から起立姿勢へ遷移を行う．
  キック動作実験では，CubiXMusashiが目標オブジェクトに対してキック動作を行う．
  これらの実験を通して，MusashiがCubiXと合体し，環境に接続したワイヤを用いて駆動することで，
  従来の筋骨格ヒューマノイドでは実現されなかった動きが達成されることを示す．
}%

\subsection{Pull-up Motion}
\switchlanguage%
{%
  \begin{figure*}[t]
    \centering
    \includegraphics[width=2.02\columnwidth]{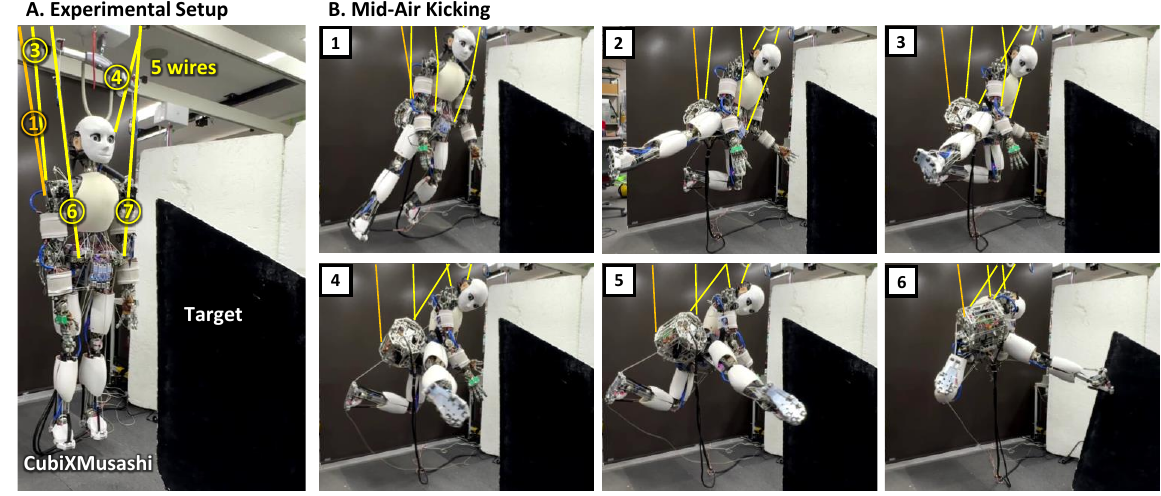}
    \vspace{-1.5ex}
    \caption{
      Mid-air kicking experiment.
      (A) Experimental setup.
      (B) Mid-air kicking.
      Using 4 wires connected to the environment, the robot assumed a kicking pose in mid-air. 
      By winding another wire connected to the environment, 
      it rotated its body in mid-air and successfully kicked a target object, as shown in these figures.
    }
    \vspace{-4ex}
    \label{fig:exp3_pic}
  \end{figure*}

  \begin{figure}[t]
    \centering
    \includegraphics[width=0.9\columnwidth]{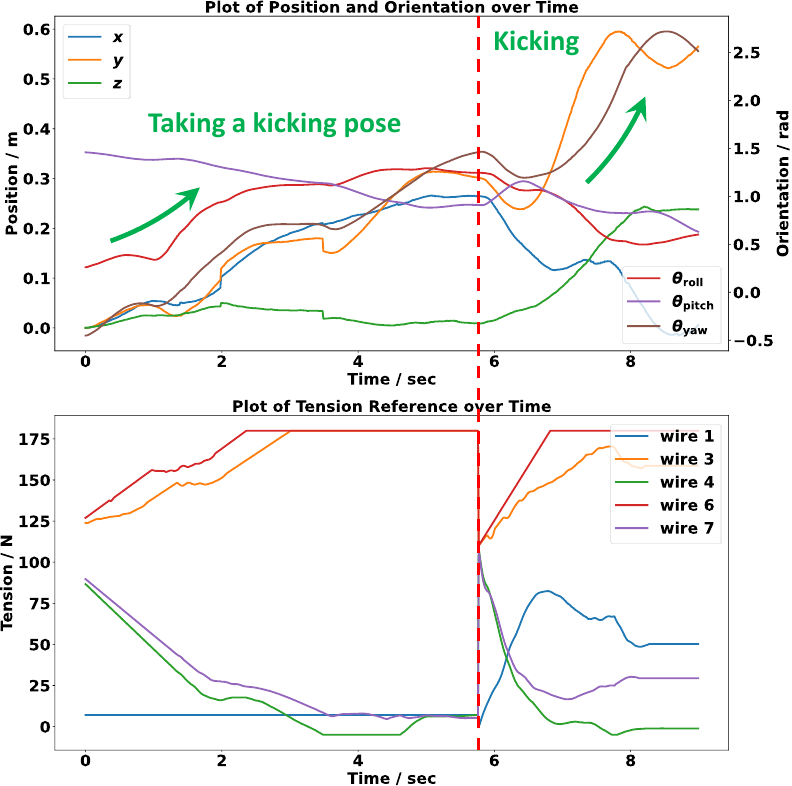}
    \vspace{-2ex}
    \caption{
      Time series data of position $x, y, z$, orientation $\theta_\mathrm{roll}, \theta_\mathrm{pitch}, \theta_\mathrm{yaw}$, 
      and target wire tension $\bm{f}^\mathrm{ref}_\mathrm{CubiX}$ during the mid-air kicking experiment.
      By controlling the lengths of 4 wires, CubiXMusashi assumes a kicking pose while rotating in the roll direction, 
      and by controlling the length of one more wire, it performs the mid-air kicking while rotating in the yaw direction.
    }
    \vspace{-4ex}
    \label{fig:exp3_data}
  \end{figure}

  In this experiment, CubiXMusashi performs a pull-up motion. 
  The experimental setup and the pull-up motion of CubiXMusashi are shown in \figref{fig:exp1_pic}, 
  and the time series data of the height position $z$ and the commanded wire tension $\bm{f}^\mathrm{ref}_\mathrm{CubiX}$ 
  during the motion are shown in \figref{fig:exp1_data}.
  As shown in \figref{fig:exp1_pic} (A), 
  CubiXMusashi connects 4 wires upward to the environment and grips the pull-up bar with both hands.
  The controlled parts of Musashi's body include the hands gripping the bar, 
  the neck, waist, and legs for maintaining body pose, while no force is generated in the arms. 
  The pull-up is executed by controlling the length of the 4 wires connected to the environment.

  As shown in \figref{fig:exp1_pic} (B), 
  CubiXMusashi achieved the pull-up motion where it grabs a bar at chest height and lifts its entire body.
  Furthermore, \figref{fig:exp1_data} indicates that the length control and gravity compensation of the entire body 
  were performed using the 4 wires connected to the environment, resulting in the elevation of CubiXMusashi's position.

  This demonstrates that CubiXMusashi successfully performed a pull-up motion, 
  lifting the entire body approximately 0.53 m while compensating for its own weight.
}%
{%
  この実験では，CubiXMusashiが懸垂動作を行う．
  実験設定とCubiXMusashiが懸垂をする様子を\figref{fig:exp1_pic}に，
  その際の高さ位置$z$と司令ワイヤ張力$\bm{f}^\mathrm{ref}_\mathrm{CubiX}$の時系列データを\figref{fig:exp1_data}に示す．
  \figref{fig:exp1_pic}の (A) の実験設定に示されるように，CubiXMusashiは4本のワイヤを上向きに環境に接続しており，
  懸垂棒を両手で握っている．
  Musashiの身体の内，制御されるのは棒を握り込んでいる手と，体幹を保つための首，腰，脚であり，腕には力を発生させていない．
  環境に接続したワイヤの長さを制御し，4本のワイヤの長さを一定量巻き取ることで，懸垂を行った．

  \figref{fig:exp1_pic}の (B) で示されるように，CubiXMusashiが胸の高さにある棒を掴んでから，
  全身を持ち上げ懸垂動作を達成できたことがわかる．
  また，\figref{fig:exp1_data}を見ると，環境に接続した4本のワイヤで全身の重力補償と長さ制御を行い，
  CubiXMusashiの位置が上昇していることがわかる．
  
  これにより，CubiXMusashiは自身の自重を補償しながら約0.53 m上昇する懸垂動作を実現したことが示される．
}%

\subsection{Rising from a Lying Pose}
\switchlanguage%
{%
  In this experiment, CubiXMusashi transitions from a lying face down on the ground to a standing pose. 
  The experimental setup and the rising motion of CubiXMusashi are shown in \figref{fig:exp2_pic}, 
  and the time series data of height position $z$, pitch rotation angle $\theta_\mathrm{pitch}$, 
  and commanded wire tension $\bm{f}^\mathrm{ref}_\mathrm{CubiX}$ during the motion are shown in \figref{fig:exp2_data}.

  As shown in \figref{fig:exp2_pic} (A), 
  CubiXMusashi connects 6 wires ($\mathrm{wire}_{2, 3, 4, 5, 6, 7}$) extending upwards from CubiX to the environment. 
  Among these 6 wires, 2 wires ($\mathrm{wire}_{6, 7}$) pass through Musashi's shoulders and connect to the environment. 
  Additionally, 
  to offset the forces attempting to bend Musashi's waist forward generated by these 2 wires ($\mathrm{wire}_{6, 7}$),
  2 wires ($\mathrm{wire}_{0, 1}$), shown in orange, are connected from CubiX to Musashi's shoulders. 
  By controlling 8 wires and Musashi's body, specifically the neck, waist, and legs, in 3 phases as follows, 
  the rising from a lying pose is achieved:
  \begin{enumerate}
    \item Lifting Phase:\\
      CubiXMusashi compensates for its own weight with 4 upward wires ($\mathrm{wire}_{2, 3, 4, 5}$), 
      controlling their lengths to lift its own body into the air. 
      During this phase, 
      the remaining 2 environment-connected wires ($\mathrm{wire}_{6, 7}$) are wound up as CubiXMusashi lifts, 
      without generating force. 
      Meanwhile, Musashi assumes a pose folding its neck and legs to facilitate rotation in the next phase. 
      The waist is controlled to prevent bending under gravity, 
      assisted by the 2 wires ($\mathrm{wire}_{0, 1}$) connected from CubiX to Musashi's shoulders, 
      which generate a greater moment for CubiXMusashi's rotation in the subsequent rotation phase.
    \item Rotation Phase:\\
      Two of the 4 weight-compensating wires ($\mathrm{wire}_{2, 5}$) are replaced 
      by 2 wires ($\mathrm{wire}_{6, 7}$) connected through the shoulders to the environment. 
      Further, by winding the 2 wires ($\mathrm{wire}_{6, 7}$) connected through the shoulders, 
      CubiXMusashi is rotated in mid-air. 
      Throughout this phase, the waist is controlled not to bend under the muscle force of Musashi 
      and the 2 wires ($\mathrm{wire}_{0, 1}$) connected from CubiX to Musashi's shoulders.
      Additionally, the neck and legs are moved into a standing pose in preparation for the next landing phase.
    \item Landing Phase:\\ 
      At this phase, CubiXMusashi's weight compensation is achieved 
      by winding the 4 wires ($\mathrm{wire}_{3, 4, 6, 7}$)
      to land CubiXMusashi on the ground with 2 legs.
  \end{enumerate}

  As shown in \figref{fig:exp2_pic} (B), 
  CubiXMusashi rotates its body in mid-air using 6 environment-connected wires and 2 internally connected wires, 
  achieving the rising from a lying pose on the ground. 
  Furthermore, \figref{fig:exp2_data} shows the control of each wire in the 3 phases, 
  confirming CubiXMusashi's actual execution of lifting, rotation, and landing.

  Thus, CubiXMusashi demonstrates the ability to rise from a lying pose on the ground to a standing pose in this experiment.
}%
{%
  \begin{figure*}[t]
    \centering
    \includegraphics[width=2.02\columnwidth]{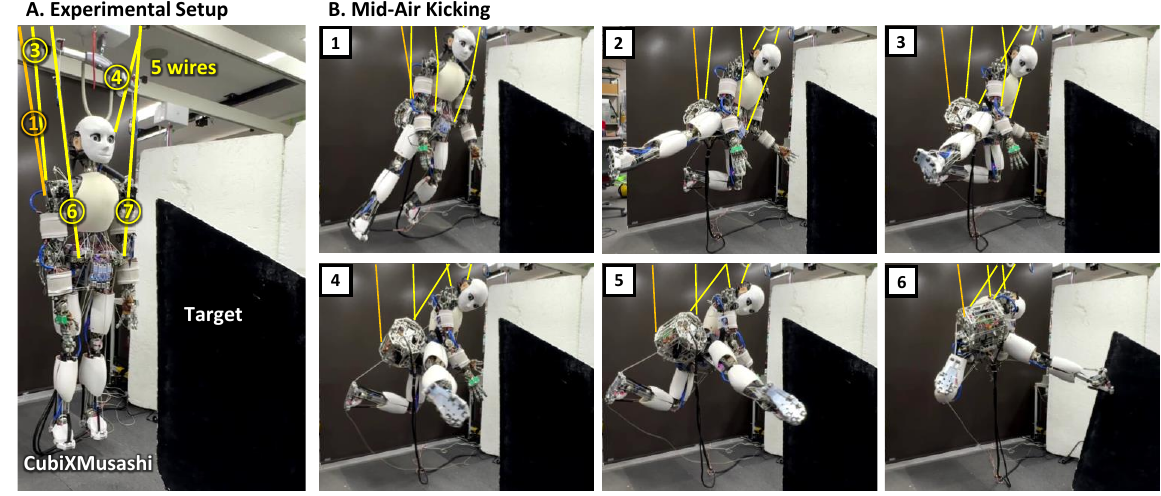}
    \vspace{-1.5ex}
    \caption{
      Mid-air kicking experiment.
      (A) Experimental setup.
      (B) Mid-air Kicking.
      Using 4 wires connected to the environment, the robot assumed a kicking pose in mid-air. 
      By winding another wire connected to the environment, 
      it rotated its body in mid-air and successfully kicked a target object, as shown in these figures.
    }
    \vspace{-2ex}
    \label{fig:exp3_pic}
  \end{figure*}

  \begin{figure}[t]
    \centering
    \includegraphics[width=1.0\columnwidth]{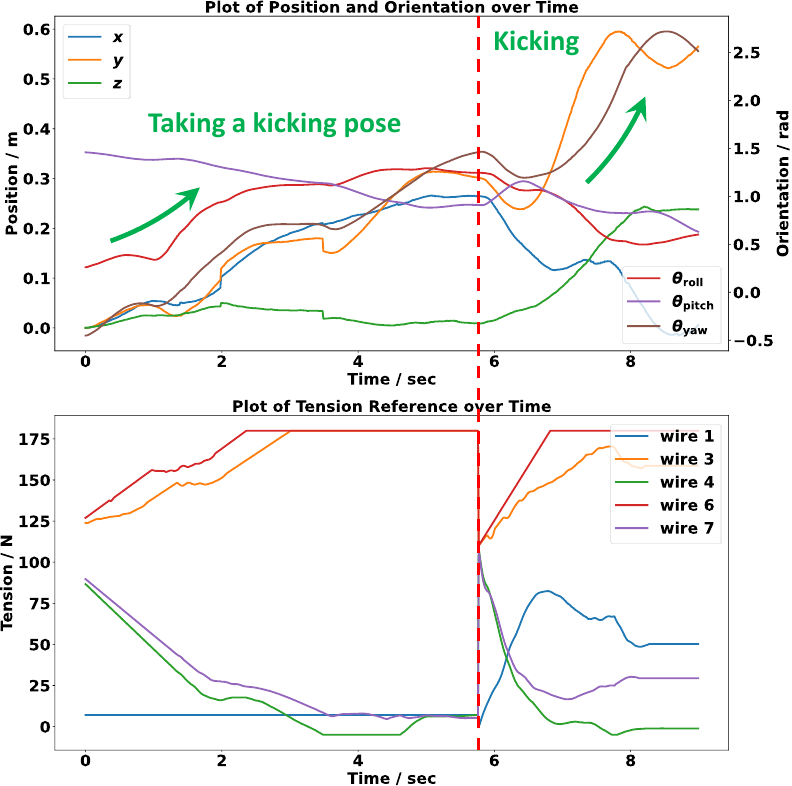}
    \vspace{-4ex}
    \caption{
      Time series data of CubiX position $x, y, z$, orientation $\theta_\mathrm{roll}, \theta_\mathrm{pitch}, \theta_\mathrm{yaw}$, 
      and target wire tension $\bm{f}^\mathrm{ref}_\mathrm{CubiX}$ during the mid-air kicking experiment.
      By controlling the length of 4 wires, 
      it shows that CubiXMusashi lifted its entire body while rotating in the roll direction to assume the kicking pose. 
      Controlling the length of another wire enabled rotation in the yaw direction to achieve the mid-air kicking.
    }
    \vspace{-5ex}
    \label{fig:exp3_data}
  \end{figure}

  この実験では，CubiXMusashiが地面にうつ伏せで横たわった状態から，起立姿勢へ起き上がる動作を行う．
  実験設定とCubiXMusashiが起き上がる様子を\figref{fig:exp2_pic}に，
  その際の高さ位置$z$とピッチ回転角度$\theta_\mathrm{pitch}$，
  司令ワイヤ張力$\bm{f}^\mathrm{ref}_\mathrm{CubiX}$の時系列データを\figref{fig:exp2_data}に示す．

  \figref{fig:exp2_pic}の (A) の実験設定に示されるように，
  CubiXMusashiはCubiXから出る8本のワイヤの内，
  黄色で示された6本のワイヤ ($\mathrm{wire}_{2, 3, 4, 5, 6, 7}$) を上向きに環境に接続している．
  その6本のワイヤの内，2本のワイヤ ($\mathrm{wire}_{6, 7}$) は，
  うつ伏せで横たわっているMusashiの肩を前側から通った上で環境に接続した．
  また，その2本のワイヤ ($\mathrm{wire}_{6, 7}$) が発生させるMusashiの腰を前側に折り曲げようとする力を相殺するために，
  オレンジ色で示された2本のワイヤ ($\mathrm{wire}_{0, 1}$) をCubiXからMusashiの肩へ接続している．
  これらの8本のワイヤとMusashiの身体の内，首，腰，脚を以下のように3段階のフェーズに分けて制御することで起き上がり動作を行う．

  \begin{enumerate}
    \item 上昇期．\\CubiXMusashiの自重を4本の上向きワイヤ ($\mathrm{wire}_{2, 3, 4, 5}$) で補償しつつ，
      この4本のワイヤの長さを一定量巻き取る長さ制御をすることで，CubiXMusashiを空中へ上昇させる．
      この時，残りの環境に接続した2本のワイヤ ($\mathrm{wire}_{6, 7}$) は，上昇に従って巻き取るだけであり，力は発生させない．
      また，Musashiは上昇中に首と脚を折りたたむ姿勢を取り，次の回転期において回転しやすい姿勢となる．
      一方で，腰はMusashiの筋肉と，CubiXからMusashiに接続された2本ワイヤ ($\mathrm{wire}_{0, 1}$) によって，
      重力に従って折れ曲がらないように制御される．
      これは，肩を通して環境に接続された2本のワイヤ ($\mathrm{wire}_{6, 7}$) が次の回転期に
      CubiXMusashiを回転させるときにより大きなモーメントを発生させるためである．
    \item 回転期．\\CubiXMusashiの自重を補償していた4本のワイヤの内，2本のワイヤを変更し，
      $\mathrm{wire}_{2, 5}$ から肩を通して環境に接続される2本のワイヤ ($\mathrm{wire}_{6, 7}$) とする．
      さらに，その肩を通して環境に接続される2本のワイヤ ($\mathrm{wire}_{6, 7}$) を一定量巻き取る長さ制御をすることで，
      空中でCubiXMusashiを回転させていく．
      この間も，回転におけるモーメントを小さくしないため，
      腰はMusashiの筋肉と，CubiXからMusashiに接続された2本ワイヤ ($\mathrm{wire}_{0, 1}$) によって，折れ曲がらないように制御される．
      また，首と脚は起立姿勢となるように動かし，次の着地期に備える．
    \item 着地期．\\この段階でCubiXMusashiの自重を補償している4本のワイヤ ($\mathrm{wire}_{3, 4, 6, 7}$) を
      一定量巻き出す長さ制御を行うことでCubiXMusashiを地面に2脚で着地させる．
  \end{enumerate}

  \figref{fig:exp2_pic}の (B) で示されるように，CubiXMusashiが地面に横たわっている状態から，
  環境に接続した6本のワイヤと身体内に接続した2本のワイヤを用いて，空中で身体を回転させ，
  起き上がり動作が達成できたことがわかる．
  また，\figref{fig:exp2_data}を見ると，上記で述べた3つの段階でそれぞれのワイヤの制御を行い，
  実際にCubiXMusashiが上昇，回転，着地を行っていることがわかる．

  これにより，CubiXMusashiは，地面にうつ伏せで横たわった状態から起立姿勢へ起き上がる動作を実現したことが示される．
}%

\subsection{Mid-Air Kicking}
\switchlanguage%
{%
  In this experiment, CubiXMusashi performs a mid-air kicking. 
  The experimental setup and the mid-air kicking of CubiXMusashi are shown in \figref{fig:exp3_pic},
  and the time series data of the CubiX center position $x, y, z$, 
  rotation angles $\theta_\mathrm{roll}, \theta_\mathrm{pitch}, \theta_\mathrm{yaw}$, 
  and the commanded wire tension $\bm{f}^\mathrm{ref}_\mathrm{CubiX}$ during the motion are shown in \figref{fig:exp3_data}.

  As shown in the experimental setup in \figref{fig:exp3_pic} (A), 
  CubiXMusashi is connected to the environment by 5 wires. 
  4 wires of them ($\mathrm{wire}{3, 4, 6, 7}$) are connected upwards, 
  and 1 wire ($\mathrm{wire}{1}$), shown in orange, is connected forwards for rotation during the kick. 
  First, while compensating for its own weight with the 4 wires connected upwards, 
  CubiXMusashi controls the lengths by winding up the 2 wires on its right side ($\mathrm{wire}{3, 6}$) 
  and unwinding the 2 wires on its left side ($\mathrm{wire}{4, 7}$) 
  to tilt its waist in the roll direction. 
  During this time, Musashi assumes a kicking pose, 
  enabling CubiXMusashi to achieve a kicking stance where the leg position is high as a result of the tilt of the entire body.
  Then, by winding the wire connected forwards ($\mathrm{wire}_{1}$), a force that rotates CubiXMusashi is generated, 
  achieving the mid-air kicking towards the target object.

  As shown in \figref{fig:exp3_pic} (B), 
  CubiXMusashi takes a kicking pose in the air and then rotates, successfully kicking the target object. 
  Additionally, \figref{fig:exp3_data} shows that, 
  by controlling each wire as described above, CubiXMusashi assumes the kicking pose in the air, rotates, and kicks.

  Thus, it is demonstrated that CubiXMusashi can achieve a kicking motion while rotating its body in the air.
}%
{%
  この実験では，CubiXMusashiがキック動作を行う．
  実験設定とCubiXMusashiがキック動作を行う様子を\figref{fig:exp3_pic}に，
  その際のCubiX中心位置$x, y, z$, 回転角度$\theta_\mathrm{roll}, \theta_\mathrm{pitch}, \theta_\mathrm{yaw}$，
  と司令ワイヤ張力$\bm{f}^\mathrm{ref}_\mathrm{CubiX}$の時系列データを\figref{fig:exp3_data}に示す．

  \figref{fig:exp3_pic}の (A) の実験設定に示されるように，
  CubiXMusashiは5本のワイヤを環境に接続しており，
  黄色で示された4本 ($\mathrm{wire}_{3, 4, 6, 7}$) は上向きに，
  オレンジ色で示された1本 ($\mathrm{wire}_{1}$) はキック時の回転のためにCubiXMusashiの前方に向かって接続している．
  まず，環境に上向きに接続した4本のワイヤで自重を補償しながら，
  CubiXMusashiから見て右側の2本のワイヤ ($\mathrm{wire}_{3, 6}$) を一定量巻き取り，
  左側の2本のワイヤ ($\mathrm{wire}_{4, 7}$) を一定量巻き出す長さ制御を行うことで，腰をロール方向に傾ける．
  この間，Musashiはキックの姿勢を取ることで，CubiXMusashiが身体全体を傾け脚の位置が高くなるキックの姿勢が達成される．
  その後，CubiXMusashiの前方に向かって接続されているワイヤ ($\mathrm{wire}_{1}$) を一定量巻き取る長さ制御をすることで，
  CubiXMusashiを回転させる力が働き，対象オブジェクトである板にキックをする動作が達成される．

  \figref{fig:exp3_pic}の (B) で示されるように，
  CubiXMusashiが空中でキックの姿勢を取り，その後に回転することで，
  対象オブジェクトである板をキックできていることがわかる．
  また，\figref{fig:exp3_data}を見ると，上記で述べた手順でそれぞれのワイヤの制御を行い，
  実際にCubiXMusashiが空中でキック姿勢を取り，回転してキックしていることがわかる．

  これにより，CubiXMusashiは，空中で身体を回転させながら行うキック動作を実現できていることが示される．
}%

\section{Discussion} \label{sec:discussion}
\switchlanguage%
{%
  We discuss the 3 experiments conducted in this study.

  In the pull-up motion experiment, 
  CubiXMusashi performed a pull-up by lifting its own weight using 4 wires connected to the environment. 
  During this process, Musashi only gripped the pull-up bar without exerting any force with its arms. 
  This demonstrates that a humanoid can generate motion using wires connected to the environment without exerting its own force. 
  Additionally, this experiment suggests the potential of using such a setup as a training platform. 
  For example, the system could gradually transfer the load of gravity compensation from the wires to the arms, 
  ultimately enabling the humanoid to achieve the desired motion independently.

  In the rising from a lying pose experiment, 
  CubiXMusashi transitioned from a lying pose on the ground to a standing pose using 8 wires. 
  Unlike the other 2 experiments, 2 wires were connected from CubiX to Musashi rather than to the environment. 
  This indicates that the performance enhancement of a wire-driven humanoid is not limited 
  to using wires connected to the environment: it can also be achieved by connecting wires within the humanoid's body 
  to reinforce the force exerted by the humanoid as needed.

  In the mid-air kicking experiment, 
  CubiXMusashi performed a kicking in mid-air using 5 wires connected to the environment. 
  While the pull-up and rising motions are challenging for musculoskeletal humanoids, 
  they are potentially achievable by general axis-driven humanoids due to their rigid bodies and superior controllability. 
  However, the mid-air rotational kicking realized in this experiment is considered difficult even for such humanoids. 
  Therefore, acquiring movements using wires connected to the environment has applicability 
  not only for musculoskeletal humanoids but also for general humanoids.

  These 3 experiments collectively demonstrate that using wires connected to the environment or the humanoid itself 
  to drive and generate motion is beneficial for humanoids.

  On the other hand, 
  the movements achieved in each experiment were realized 
  by following manually generated target wire lengths $\bm{l}^\mathrm{ref}_\mathrm{CubiX}$. 
  CubiXMusashi did not autonomously understand its positional relationship with the environment or the arrangement of the wires, 
  nor did it consider how to utilize the wires connected to the environment. 
  To truly remove the limitations on humanoid performance, 
  it is necessary for the humanoid to recognize its surrounding environment, 
  design and execute the wire arrangements considering the desired movements, and utilize them effectively.
}%
{%
  本研究で行った3つの実験についてそれぞれ考察する．

  懸垂動作実験では，環境に接続した4本のワイヤを用いて自重を持ち上げることにより，懸垂を行った．
  この際，Musashiは懸垂棒を握りしめているだけであり，腕に力をかけて駆動することはしていない．
  ヒューマノイドが力を出さずとも環境に接続したワイヤを用いて運動を生成できることを示すと同時に，
  例えば徐々に環境に接続したワイヤが負担する重力補償を腕へと移していき，
  最終的にはヒューマノイドのみで所望の動作を獲得していくというような
  運動訓練プラットフォームとしての機能も考えられるかもしれない．

  起き上がり動作実験では，8本のワイヤを用いて，地面にうつ伏せで横たわった状態から起立姿勢への起き上がりを行った．
  用いた8本のワイヤの内，2本のワイヤは，環境に接続せず，CubiXからMusashiへ接続されたことが他の2つの実験と異なる．
  ワイヤ駆動を用いたヒューマノイドの性能向上は，そのワイヤを環境に接続して利用することに留まらず，
  ヒューマノイドの体内に接続し，ヒューマノイドが発揮する力を目的に合わせて強化することでも達成されることが示されている．

  キック動作実験では，環境に接続した5本のワイヤを用いて空中でのキック動作を行った．
  懸垂動作や起き上がり動作は，筋骨格ヒューマノイドにとっては難易度が高い動作であるが，
  身体が剛であり制御性に優れる一般的な軸駆動のヒューマノイドにとっては実現できる可能性のある動作である．
  しかし，この実験で実現した空中で回転するキック動作は，そのようなヒューマノイドにとっても実現し難い運動であると考えられる．
  したがって，環境に接続したワイヤによる駆動を活かした動作の獲得は，
  筋骨格ヒューマノイドに限らず一般的なヒューマノイドにとっても適用価値があると考えられる．

  このように，これら3つの実験を通して，環境や自分自身に接続したワイヤを用いて駆動し，運動を生成することが，
  ヒューマノイドにとって有用であることが示された．

  一方で，各実験で実現した動きは，
  どれも手作業で生成した目標ワイヤ長さ$\bm{l}^\mathrm{ref}_\mathrm{CubiX}$に追従することで実現されており，
  CubiXMusashiが，環境との位置関係やワイヤの配置を自ら把握し，環境に接続されたワイヤをどのように利用するかを考えているわけではない．
  ヒューマノイドのパフォーマンスにおける制限を真に取り払うには，
  ヒューマノイドが周りの環境を認識し，所望の運動との兼ね合いを考えながらワイヤの配置を設計，実行し，
  利用できるようになることが求められる．
}%

\section{Conclusion} \label{sec:conclusion}
\switchlanguage%
{%
  In this study, we demonstrated that CubiXMusashi,
  a fusion of the wire-driven robot CubiX capable of connecting to the environment and the musculoskeletal humanoid Musashi, 
  can generate motion by winding wires connected to the environment or itself. 
  By integrating CubiX with Musashi and connecting up to 8 wires from the body to the environment for actuation, 
  we achieved motions that were previously unattainable in musculoskeletal humanoids, 
  which are characterized by their flexible and complex bodies with high degrees of freedom.

  In the experiments, 
  we successfully performed motions that are challenging for Musashi alone, 
  such as pull-up motion lifting the entire body approximately 0.53m,
  and rising from a lying pose
  to a standing pose. 
  Additionally, we confirmed the realization of high-difficulty movements, even for general humanoids, 
  such as kicking a target object while rotating the entire body in mid-air.

  These results demonstrate that both musculoskeletal humanoids and general humanoids can acquire new capabilities 
  and movements that were previously unattainable by utilizing wires connected to the environment or to themselves for actuation.
}%
{%
  本研究では，筋骨格ヒューマノイドであるMusashiに，環境接続可能なワイヤ駆動ロボットであるCubiXを合体させた
  CubiXMusashiが，環境や自分自身に接続したワイヤを巻き取り，駆動することによって運動生成を行った．
  MusashiにCubiXを合体させ，身体から環境へ最大8本のワイヤを接続して駆動することで，
  筋骨格ヒューマノイドに見られる柔軟で複雑かつ高制御自由度な身体において，
  これまで実現されなかった運動を実現した．
  実験では，懸垂棒を掴み，全身を約0.53 m 持ち上げる懸垂動作や，
  地面に横たわった状態から起立姿勢へ起き上がる動作など，
  Musashi単体では実現の難しい動作に成功した．
  また，空中で全身を回転させながら対象物をキックする動作といった，
  一般的なヒューマノイドにとっても実現難易度が高い動作の実現が確認された．
  以上より，筋骨格ヒューマノイドや，一般にヒューマノイドが，
  環境や自身に接続したワイヤを用いて駆動することによって，
  これまでに実現できなかった新たな能力や動作を獲得できることが示された．
}%

{
  \bibliographystyle{IEEEtran}
  \bibliography{bib}
}

\end{document}